\pdfoutput=1

\documentclass[11pt]{article}

\usepackage{acl}

\usepackage{times}
\usepackage{latexsym}

\usepackage[T1]{fontenc}

\usepackage[utf8]{inputenc}

\usepackage{microtype}

\usepackage{amsmath}
\usepackage{cleveref}
\crefname{section}{§}{§§}
\usepackage{tabularx}
\usepackage{booktabs}
\usepackage{graphicx}
\usepackage{enumitem}
\usepackage{xcolor}
\usepackage[belowskip=0pt,aboveskip=0pt]{caption}
\usepackage{multirow}
\usepackage{MnSymbol}
\usepackage{subcaption}
\usepackage{xspace}

\title{Long Context vs. RAG for LLMs: An Evaluation and Revisits}

\author{
 Xinze~Li$^{1}$, Yixin~Cao$^{2\dag}$, Yubo~Ma$^{1}$, Aixin Sun$^{1\dag}$ \\ 
 \\
 $^1$ S-Lab, Nanyang Technological University \\
 $^2$ School of Computer Science, Fudan University\\
\texttt{\{xinze002, yubo001\}@e.ntu.edu.sg} \hspace{1cm}
\texttt{axsun@ntu.edu.sg} \\
\texttt{yxcao@fudan.edu.cn}\\
}

\begin{document}
\maketitle
\begin{abstract}
Extending context windows (\textit{i.e.,} Long Context, LC) and using retrievers to selectively access relevant information (\textit{i.e.,} Retrieval-Augmented Generation, RAG) are the two main strategies to enable LLMs to incorporate extremely long external contexts. This paper revisits recent studies on this topic, highlighting their key insights and discrepancies. We then provide a more comprehensive evaluation by filtering out questions answerable without external context, identifying the most effective retrieval methods, and expanding the datasets. We show that LC generally outperforms RAG in question-answering benchmarks, especially for Wikipedia-based questions. Summarization-based retrieval performs comparably to LC, while chunk-based retrieval lags behind. However, RAG has advantages in dialogue-based and general question queries. These insights underscore the trade-offs between RAG and LC strategies, offering guidance for future optimization of LLMs with external knowledge sources. We also provide an in-depth discussion on this topic, highlighting the overlooked importance of context relevance in existing studies.
\end{abstract}

\maketitle
\begin{figure*}[t]
    \centering
    \begin{subfigure}{\linewidth}
    \includegraphics[width=\linewidth]{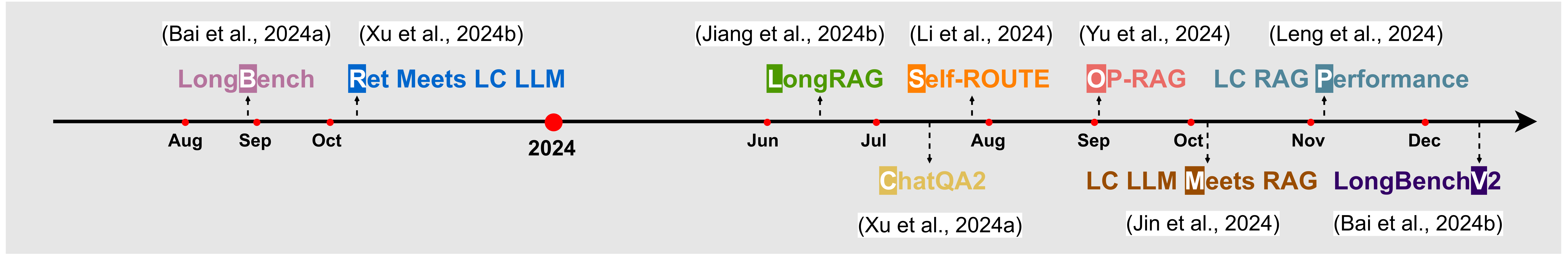}
        \caption{Related work on LC and RAG, each paper is labeled by a char and one color. For instance, green and "L" represent "LongRAG".}
        \label{fig:subgraph1}
    \end{subfigure}
    \vspace{1em} 
    \begin{subfigure}{\linewidth}
             \includegraphics[width=\linewidth]{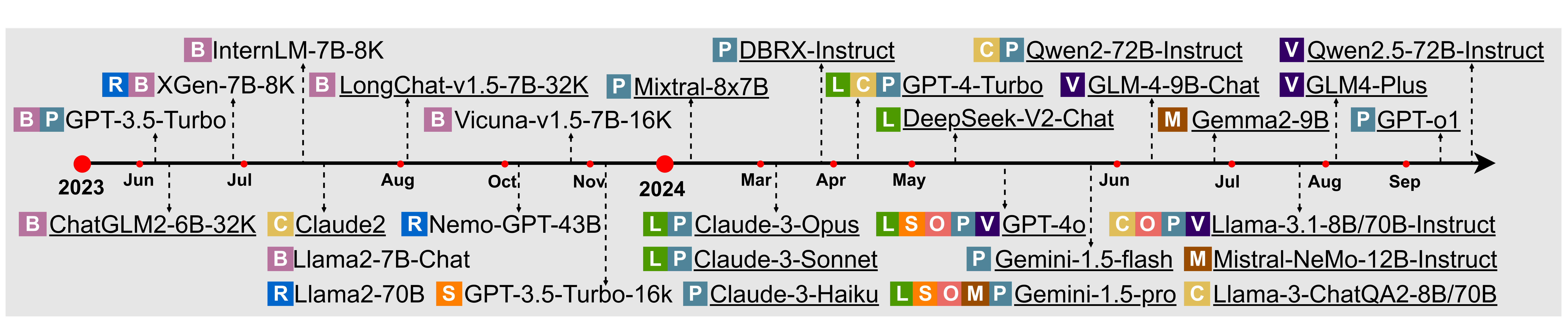}
        \caption{Chronological progress of key LLMs from 2023 to 2024. We focus on the models that publications in \ref{fig:subgraph1} use. We \underline{underline} the models that support context window length of $\geq 32K$.}
        \label{fig:subgraph2}
    \end{subfigure}
    \vspace{1em} 
    \begin{subfigure}{\linewidth}
       
        \includegraphics[width=\linewidth]{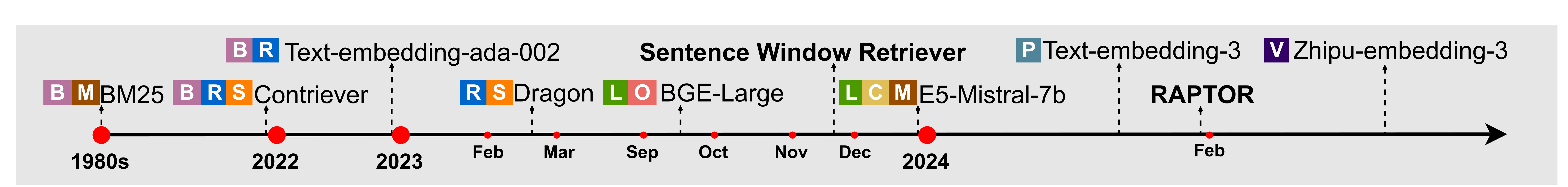}
        \caption{History of frequently used retrievers from the 1980s until 2024. We \textbf{bold} the retrievers that no existing publications in \ref{fig:subgraph1} uses. }
        \label{fig:subgraph3}
    \end{subfigure}
    \caption{Chronological overview of the development of RAG and LC. The Sub-graphs respectively illustrate the timelines for (a) publications related to LC and RAG, (b) long-context models, and (c) retrievers. We label before each model and retriever with the char and color block representing the publication that uses it.}
    \label{fig:timelines}
\end{figure*}

\section{Introduction}

Large Language Models (LLMs)~\cite{brown2020language} have demonstrated strong zero/few-shot capabilities in open-ended question answering~\cite{yang-etal-2019-end-end}. However, they face challenges such as hallucinations~\cite{shuster-etal-2021-retrieval-augmentation, ji2023survey}, lacking real-time information and domain-specific knowledge~\cite{su2024dragin,zhang2024raft}, among others. A common solution is to enhance LLMs with external memory to provide reliable and up-to-date data sources. Yet, incorporating additional content is constrained by the limited context window of LLMs. To address this, two main approaches are adopted: (i) building models with long context windows to read in more information (\textbf{LC})~\cite{fei2023extending, chen2023extending, wang2024beyond}, and (ii) employing retrievers to include text segments relevant to the query (\textbf{RAG})~\cite{jiang-etal-2023-active, asai2023self, gao2023retrieval}.

As shown by the timeline in Figure~\ref{fig:timelines}a, there is a clear trend toward developing models that handle longer context windows and combining LC with RAG methods. The chronological overview of related studies highlights an increasing focus on both LC and RAG since mid-2023, as evidenced by a growing number of publications aimed at optimizing the efficient retrieval, and utilization of long contexts. The development of models supporting longer context windows underscores the growing importance of handling extensive inputs effectively.

Despite the broad consensus regarding the importance of LC and RAG, there remain disagreements and contradictory insights from different studies, summarized in Table~\ref{tab:insights}. For example, while several studies agree on the effectiveness of combining LC and RAG~\cite{xu2023retrieval,jiang2024longrag}, others suggest that combining may not be beneficial~\cite{bai-etal-2024-longbench,jin2024long}. Moreover,  conflicting conclusions are reported regarding the benefits of RAG versus LC. Some papers find RAG advantageous in certain contexts~\cite{xu2024chatqa,yu2024defense}, while others highlight superior results from LC~\cite{li2024retrieval,xu2023retrieval}. These divergent insights showcase the complexity and ongoing debates in the field, suggesting that optimal strategies may vary depending on specific model architectures and benchmark conditions.

To explore the underlying reasons, we conduct an in-depth investigation into the conditions that lead to disagreements among existing studies. During this process, we also identify key aspects that may have been overlooked in earlier research. Specifically, we revisit the evaluation process and implement the following changes. First, we filter out questions from existing datasets that can be correctly answered without external context, removing biases from the parametric knowledge of LLMs and focusing on questions requiring external knowledge. Second, we evaluate retrieval methods and baselines on a smaller filtered dataset (1,000+ questions) from 12 QA datasets to identify the best retriever. Third, we expand the dataset size by approximately 10 times by collecting additional data from the original sources of the 12 datasets\footnote{The experiment code and expanded datasets are available at \url{https://github.com/lixinze777/LC_VS_RAG}}. Lastly, we compare the answers produced by the two settings, \textit{i.e.}, LC and RAG, and conduct an in-depth analysis. Our results are based on the expanded dataset using the long-context setting and the best retrieval method identified earlier.

Our key contributions in this paper are as follows: (i) Providing a comprehensive survey of existing studies on LC and RAG, analyzing their implementations and key insights. (ii) Proposing a fair and systematic evaluation framework, and performing detailed analyses to understand the strengths and limitations of LC and RAG. (iii) Discussing challenges for comparing and combining LC and RAG, reflecting on the key points that researchers tend to overlook in this field.
Evaluation results indicate that LC models generally outperform RAG when processing self-contained information like stories, while RAG excels at handling fragmented information, particularly in dialogue-based contexts. These experiments deepen our understanding of the strengths and limitations of LC and RAG, offering valuable insights into optimizing retrieval strategies and effectively integrating these approaches to enhance performance in open-domain question answering. These findings also based on a systematic survey of existing studies on this topic (see~\cref{sec:related work}). Additionally, we discuss key aspects of comparing LC and RAG in~\cref{sec:discussion}, highlighting areas that have been underexplored in prior research.
\section{Related Work}
\label{sec:related work}

Our primary focus is to evaluate and compare LC and RAG. To this end, we review papers with a similar focus, and provide a detailed analysis of the retrievers and long-context settings they employ.

\subsection{Retrievers}
Retrievers, as fundamental components of RAG pipelines, focus on identifying and extracting contextually relevant segments of documents. We categorize retrieval strategies into three main approaches: \textit{chunk-based retrieval}, which splits documents into smaller segments and then retrieves those most relevant to a query; \textit{index-based retrieval}, which builds specialized index structures to guide efficient and context-rich lookups; and \textit{summarization-based retrieval}, which leverages hierarchical summaries to capture a document’s key information at various levels of abstraction.

\paragraph{Chunk-based Retrieval} can be broadly categorized into sparse retrievers and dense retrievers. Sparse retrievers, such as the classic BM25~\cite{robertson2009probabilistic}, operate on term frequency-based representations of text and rank chunks based on a similarity function, leveraging exact matches and term weighting. With the advent of word embeddings, dense retrievers have gained prominence. These models encode both queries and document chunks into dense vector representations and calculate relevance using similarity metrics, such as cosine similarity.  

Since text similarity is often defined by measuring the distance between embeddings, the quality of these embeddings is particularly important. 
Contriever~\cite{izacard2021unsupervised} leverages contrastive learning for training without supervision. By generating synthetic queries and pre-training on unlabeled data, Contriever provides robust retrieval capabilities especially in cross-lingual applications. 
On a larger scale, BGE-Large~\cite{bge_embedding} employs diverse datasets and sophisticated training methods to outperform previous models on comprehensive benchmarks such as C-MTEB. 
E5Mistral-7b~\cite{wang2024improvingtextembeddingslarge} combines open-source, decoder-only LLMs with synthetic data generation pipelines. With minimal human annotations, the fine-tuning achieves SOTA performance on BEIR and MTEB.
Dragon~\cite{lin-etal-2023-train} also employs data augmentation, including cropping and generative queries, and integrates labels from multiple retrieval sources. This strategy ensures its effectiveness without increasing model complexity.
Another method of learning high-quality embeddings is through strong generalization ability from LLMs. For instance, OpenAI embeddings draw upon the GPT-3.5/4 family while Zhipu-embedding-3 leverages the GLM family~\cite{glm2024chatglm}.

\paragraph{Index-based Retrieval} requires pre-processing on the documents with more complicated data structures~\cite{gupta2018document}.  
With the development of LLM, Llama-Index~\cite{Liu_LlamaIndex_2022} was proposed to facilitate interaction between the model and documents more conveniently. The index provides a flexible interface to construct various data structures, known as ``indices'' that store, organize, and facilitate quick retrieval of context. Once created, these indices can be efficiently queried, guiding the LLM to the most relevant information, improving the accuracy of responses. Some classic indexing methods include tree index which constructs a hierarchical tree from nodes, and knowledge graph index, which builds a knowledge graph with labeled nodes and relationships.

\paragraph{Summarization-based Retrieval} is built on top of chunk- and index-based approaches. It provides comprehensive summaries for key points in a document. These summaries available for retrieval.

RAPTOR~\cite{sarthi2024raptor} improves retrieval by generating recursive summaries of text chunks organized in a tree structure. Instead of retrieving short, contiguous text snippets, RAPTOR clusters text segments, summarizes them at various levels, and forms a hierarchical tree that represents the document's content at different levels of abstraction. This allows retrieval models to extract context at varying levels of detail, improving the ability to handle complex questions that require synthesizing information from multiple parts of the document. Such a summarization-based retrieval method enhances retrieval accuracy for tasks requiring long-range or multi-step reasoning.

\subsection{Long-Context LLMs}
Many research efforts focus on extending input and output windows to accommodate more context (see Figure~\ref{fig:subgraph2}), enabling applications such as extended dialogues, large document processing, and complex multimodal tasks. Thus, our analysis focuses on two dimensions: the model capabilities and the context length they can reach.

\paragraph{Model Ability.}
While most of the models discussed here excel at understanding long documents, many emphasize specialized capabilities. ChatGLM2-6B-32K~\cite{glm2024chatglm} employs Multi-Query Attention to achieve high reasoning efficiency with low memory usage, making it suitable for tasks requiring deep reasoning. XGen-7B-8K~\cite{Xgen} enhances long-context conversational understanding and text summarization, enabling coherent and contextually rich dialogues. InternLM-7B-8k~\cite{cai2024internlm2} is optimized for knowledge understanding, reading comprehension, and multilingual translation, supporting diverse linguistic applications. Models like DeepSeek-V2-Chat~\cite{deepseekv2}, Qwen2-72B-Instruct~\cite{yang2024qwen2technicalreport}, Qwen2.5-72B-Instruct~\cite{qwen2.5}, Mixtral-7x8b~\cite{jiang2024mixtralexperts}, and DBRX-Instruct excel in mathematical computations, logical reasoning, and coding, demonstrating strong performance in technical and analytical tasks.

Additionally, Claude-3-Opus, Sonnet, Haiku, Gemini-1.5-flash, and Gemini-1.5-pro~\cite{geminiteam2024gemini15unlockingmultimodal} incorporate multi-modal capabilities, effectively handling both textual and visual information. GLM-4-9B-Chat~\cite{glm2024chatglm}, Mistral-12b-Instruct, and Llama-3.1-Instruct~\cite{dubey2024llama3herdmodels} offer robust multilingual abilities, strong instruction-following and multi-turn dialogue capabilities, increasing their utility in a wide range of conversational scenarios. Finally, Claude-2 is notable for low hallucination rate when processing extra-long documents, ensuring high accuracy and reliability in information retrieval and synthesis.

\paragraph{Context Length.}
As shown in Figure~\ref{fig:subgraph2}, there is a clear trend of increasing context length in newly released models. Following the categorization approach proposed by ChatQA2~\cite{xu2024chatqa}, we classify these models into three categories based on their supported context windows: short (up to 4K), long (up to 32K), and ultra-long (more than 32K) context models.

\textit{Short context} models, such as Llama2-70B and llama2-7B-chat-4k~\cite{touvron2023llama2openfoundation}, support up to 4K tokens and are typically employed as baselines for retrieval and standard conversational tasks.
\textit{Long context} models, including XGen-7B-8K\cite{Xgen}, InternLM-7B-8k\cite{cai2024internlm2}, Mixtral-7x8b~\cite{jiang2024mixtralexperts}, DBRX-Instruct and Gemma2-9B~\cite{gemmateam2024gemmaopenmodelsbased}, offer context windows ranging from 8K to 32K tokens. These are ideal for extended conversations, comprehensive text analysis, and detailed summarization tasks.
\textit{Ultra-long context} models extend beyond 32K tokens. For example, Claude-2 provides a 100K token window, while Claude-3-Opus, Sonnet, and Haiku handle up to 200K tokens. GPT-4-Turbo\cite{openai2024gpt4technicalreport}, GPT-4o, and GPT-o1 all support 128K tokens, as do DeepSeek-V2-Chat\cite{deepseekv2}, Qwen2-72B-Instruct\cite{yang2024qwen2technicalreport}, Qwen2.5-72B-Instruct~\cite{qwen2.5}, GLM-4-9B-Chat~\cite{glm2024chatglm}, GLM-4-Plus, Mistral-12b-Instruct, and Llama-3.1-Instruct. Notably, Gemini-1.5-flahs and Gemini-1.5-pro\cite{geminiteam2024gemini15unlockingmultimodal} both support up to an unprecedented 10M tokens. These ultra long-context models enable the processing of exceptionally large documents, complex multimodal tasks, and extensive multi-turn dialogues.

\begin{table*}
\centering
\small  
\begin{tabularx}{0.85\textwidth}{l|c|l}
\toprule
\textbf{Paper} & \textbf{Type} & \textbf{Findings} \\ \midrule
\textbf{LongBench} (\textbf{B}) & $\bullet$ & Retrieval helps 4k model, but not 16k/32k models.  \\
\cite{bai-etal-2024-longbench}& $+$ & Models benefit from continuous training on long contexts.\\ 
& $+$ & Splitting context into shorter and more chunks is better. \\ \midrule
\textbf{Ret-LC LLM} (\textbf{R}) & $\star$ & LC is better for multi-hop benchmarks than 4k RAG.  \\
\cite{xu2023retrieval}& $\circ$ & RAG improves on 70B/43B models on all context lengths. \\ 
& $+$ & For LC model, best results are obtained from top-5 or top-10. \\ \midrule
\textbf{LongRAG} (\textbf{L}) & $\circ$ & Retrieval benefits from long retrieval units. \\ 
\cite{jiang2024longrag}&&\\\midrule
\textbf{ChatQA2} (\textbf{C}) & $\smallstar$ & For sequence lengths up to 32K, RAG outperforms LC. \\
\cite{xu2024chatqa}& $\circ$ & From 3K to 24K, greater context window benefits RAG. \\ \midrule
\textbf{Self-ROUTE} (\textbf{S}) & $\star$ & LC consistently outperforms RAG, but RAG has lower cost.   \\ 
\cite{li2024retrieval}&&\\\midrule
\textbf{OP-RAG} (\textbf{O}) & $\smallstar$ & Efficient retrieval can outperform brute-force LC. \\
\cite{yu2024defense}& $+$ & Too many chunks in RAG harms performance. \\
& $+$ & Preserving the original order is better than ordering by score. \\  \midrule
\textbf{LC LLM-RAG} (\textbf{M}) & $\bullet$ & Retrieve more passages first improves performance then drops. \\
\cite{jin2024long}& $+$ & Ordering higher score information to front and back helps. \\ \midrule
\textbf{LC RAG} & $\circ$ & Most close models' RAG improves up to 100k tokens. \\
\textbf{Performance} (\textbf{P})& $\bullet$ & Most open models' RAG peak at 16k-32k then performance drops. \\
\cite{leng2024longcontextragperformance}& & \\\midrule
\textbf{LongBench v2} (\textbf{V}) & $\smallstar$ & GPT-4o performs better at 128k without RAG. \\
\cite{bai2024longbenchv2deeperunderstanding} & $\circ$ & GPT-4o performance keeps increasing to 128k RAG context. \\
& $\bullet$ & Qwen2.5 \& GLM-4-Plus drop with >32k RAG contexts. \\ \bottomrule
\end{tabularx}
\caption{Important findings from existing studies that compare or combine LC with RAG (label in brackets). We group the insights into three categories: 
1) General strategies that improve performance marked by $+$.
2) Combining LC and RAG, where $\circ$ indicates combining is good, and $\bullet$ for combining is not helpful, and
3) Comparing LC and RAG, where $\smallstar$ indicates RAG outperforms LC, and $\star$ for LC outperforms RAG. 
}
\label{tab:insights}
\end{table*}

\subsection{Comparing \& Combining LC and RAG}
Since the increase in LLMs’ context window lengths, some models can contain the entire document, reducing the need to retrieve on documents. Hence, more studies have begun comparing the performance of long-context LLMs and RAG, as well as investigating ways to combine them.
LongBench~\cite{bai-etal-2024-longbench} conducts early comparison experiments on a 4K model with RAG and a 32K model. \citet{xu2023retrieval} systematically compare LC LLMs and RAG, and proposes their combination. LongRAG~\cite{jiang2024longrag} introduces long retrievers and long readers, a successful application of long retrieval units to RAG. ChatQA2~\cite{xu2024chatqa} instruction-tunes long-context LLMs to a 128K context window and tests their ability with long-context retrievers. Self-ROUTE~\cite{li2024retrieval} enables the model to select either RAG or LC based on self-reflection to reduce costs. OP-RAG~\cite{yu2024defense} preserves the original order of retrieved chunks, and LC LLM meets RAG~\cite{jin2024long} investigates long-context LLMs in RAG systems, proposing retrieval reordering methods. LC RAG Performance of LLM~\cite{leng2024longcontextragperformance} evaluates the effectiveness of RAG on long-context LLMs across context lengths from 2K to 2M tokens.
Very recently, LongBench is updated to LongBench V2~\cite{bai2024longbenchv2deeperunderstanding}, which tests LLMs on long context comprehension and reasoning with a more realistic and challenging setting.

We summarize the key insights from these papers into three categories: (1) general insights such as chunking strategies, (2) combining the two strategies, and (3) comparing the performance between LC and RAG (see Table~\ref{tab:insights}).   

Some papers reach consensus on chunking strategy that, retrieval units should be longer~\cite{jiang2024longrag} and the number of chunks should be kept low~\cite{yu2024defense}. According to \cite{xu2023retrieval}, selecting the top 5 to 10 chunks typically yields strong performance, while retrieving more than 20 chunks leads to diminished results. LongBench~\cite{bai-etal-2024-longbench} presents a different finding, suggesting that splitting a long context into shorter and more numerous chunks is better. However, at the time of its publication, LLMs generally exhibited weaker long-context capabilities, and the study did not incorporate very long retrieval units (>1000 tokens). Consequently, LongBench’s findings are not at odds with the broader consensus. 

Nonetheless, these papers present disagreement regarding performance of retrieval on long-context LLMs. For instance, LongBench~\cite{bai-etal-2024-longbench} finds that retrieval helps short-context models but not 7B long-context models. In contrast, \citet{xu2023retrieval} suggest that RAG improves 70B models across all context lengths, attributing the discrepancy to the difference between model sizes. Similarly, ChatQA2~\cite{xu2024chatqa} observes that increasing the context window from 3K to 24K tokens consistently benefits RAG. Notably, LongBench V2~\cite{bai2024longbenchv2deeperunderstanding} shows that GPT-4o continues to improve in RAG performance even at 128K input, whereas Qwen2.5 and GLM-4-Plus show performance deterioration beyond 32K input. The observations align with findings from ~\cite{leng2024longcontextragperformance} that RAG for close-source models can improve up to 100K input, whereas performance for some open-source models peaks around 16K tokens. Hence, the varying behaviors might be due to different model size and architecture. 

There are even greater discrepancies in the direct comparisons between the two methods. \citet{xu2023retrieval} claims that long-context models outperform retrieval with short-context models in multi-hop benchmarks. In contrast, ChatQA2~\cite{xu2024chatqa} finds that RAG can outperform LC if a sufficient number of top-k chunks are used. Self-ROUTE~\cite{li2024retrieval} fully supports LC, arguing that it outperforms RAG in all benchmarks. Meanwhile, OP-RAG~\cite{yu2024defense} defends RAG, demonstrating that efficient retrieval strategies can outperform a brute-force approach of processing extremely long contexts. 

The reasons for the differences among these studies are manifold. For instance, There are three categories of retrieval methods (\textit{i.e.,} chunk-based, index-based, and summarization-based retrieval), but current studies rely predominantly on chunk-based retrieval, leaving room for further optimization. Additionally, evaluation scores often represent weighted averages across different datasets. Because each dataset has distinct characteristics, placing more emphasis on one dataset and less on another can alter the final results. Finally, most existing studies use only a few datasets with around 200 questions each. This small sample size creates greater room for variability and reduces the general reliability of these findings.
\begin{table*}[htbp]\small\resizebox{\textwidth}{!}{\begin{tabular*}{\textwidth}{l|c|c|c|r|l||rrrc}\toprule
\textbf{Dataset} & \textbf{T} & \textbf{Doc} & \textbf{Source} & \textbf{Avg Len} & \textbf{Used by Papers} & \textbf{\# Q} & \textbf{\# Kept} & \textbf{\% Kept} &\textbf{Mode} \\ \midrule
NQ & K & multi & Wikipedia & 18,164.7 & M, P & 109 & 22 & 20 & Open\\
Coursera & K & multi & Coursera & 7,934.3 & NIL (L-eval) & 172 & 54 & 32 & MCQ\\
NovelQA & C & single & books & 67,000.0 & NIL (NovelQA) & 210 & 109 & 52 & MCQ\\
2WikiMHQA & R & multi & Wikipedia & 7,191.3 & B, S, M & 300 & 152 & 51 & Open\\
HotpotQA & R & multi & Wikipedia & 10,602.7 & B, R, L, C, S, M & 200 & 93 & 47 & Open\\
MuSiQue & R & multi & Wikipedia & 12,974.3 & B, R, C, S & 200 & 140 & 70 & Open \\
MultiFieldQA & C & single & papers, reports & 5,706.1 & B, R, L, C, S & 150 & 121 & 81 & Open\\
NarrativeQA & C & single & books, films & 25,274.2 & B, R, S & 200 & 171 & 86 & Open\\
QASPER & C & single & papers & 5,350.3 & B, R, C & 224 & 221 & 99 & Open\\
QuALTY & C & single & stories & 5,089.2 & R, C & 202 & 202 & 100 & MCQ\\
TOEFL-QA & C & single & exams & 729.1 & NIL (L-eval) & 121 & 121 & 100 & MCQ\\
MultiDoc2Dial & C & multi & dialogue & 3,076.9 & NIL (L-eval) & 158 & 158 & 100 & Open\\
\bottomrule
\end{tabular*}} \caption{Overview of the original datasets (i.e., the pre-expanded \textit{sample question set}) and their characteristics. The column ``T'' represents dataset type with values ``K'' for ``Knowledge'', ``R'' for ``reasoning'', and ``C'' for ``reading comprehension''. 
For each dataset, we report the existing papers (with the label) about LC \& RAG that use it. If no paper has used it, we report its source like L-eval~\cite{an-etal-2024-l}.
We also report number of questions in each set (\# Q), number and percentage of questions retained after filtering (\# Kept and \% Kept) out questions needing no context, and mode of question.} \label{tab:benchmark_overview}
\end{table*}

\section{Question Filtering and Expansion}
To ensure a fair and comprehensive comparison, we curate our evaluation dataset based on existing datasets, and apply necessary filtering (\cref{subsec: dataset_filtering}) and augmentation (\cref{subsec: data_augmentation}). 
We select 12 long-context QA datasets frequently used in studies comparing LC and RAG: Natural Questions~\cite{kwiatkowski-etal-2019-natural}, 2WikiMultihopQA~\cite{ho-etal-2020-constructing}, HotpotQA~\cite{yang-etal-2018-hotpotqa}, MuSiQue~\cite{trivedii-etal-2022-musique}, MultiFieldQA~\cite{bai-etal-2024-longbench}, NarrativeQA~\cite{kocisky-etal-2018-narrativeqa}, QASPER~\cite{dasigi-etal-2021-dataset}, QuALTY~\cite{pang-etal-2022-quality}, Coursera, TOEFL-QA, and MultiDoc2Dial~\cite{an-etal-2024-l}. We also include the NovelQA~\cite{wang2024novelqa} dataset, a high-quality, human-annotated resource derived from long-form novels. We present an overview of these datasets in Table \ref{tab:benchmark_overview}, including their type, context type (single-doc or multi-doc), context source, average context length, and representative studies that have utilized each dataset.

\subsection{Question Filtering}\label{subsec: dataset_filtering}
Given the strong capabilities of modern LLMs, many questions can be directly answered based on knowledge encoded in their parameters~\cite{basmov2024llms}, reducing the need for external context in some cases. However, certain queries, such as those related to private conversations, will always require additional context.
To determine which approach more effectively enhances an LLM’s performance with long documents, we filter the datasets to include only questions that the LLM cannot answer correctly without external context. This ensures that any correct answers obtained subsequently must rely on external knowledge rather than the model’s built-in knowledge.

For our implementation, we use GPT-4o for question filtering due to its strong capabilities. We employ a strict exact-match scoring metric to ensure that the model not only provides the correct answer but also demonstrates a complete understanding of the required information.

\subsection{Question (and Context) Expansion}\label{subsec: data_augmentation}
RAG and LC produce identical answers for about 60\% of the questions 
in existing evaluations~\cite{li2024retrieval}, leaving relatively few questions to help us understand the differences between the two. To ensure robust statistical significance, we expand the dataset size to approximately 20,000 questions by collecting additional samples.

To maintain a similar distribution as the original datasets, we follow two principles during data collection. First, we collect questions only from the original source of each dataset, avoiding artificially generated or LLM-augmented questions. Second, we add distracting passages to the original context for each question to extend the context length, following the implementation described in LongBench. For NovelQA, we use all its available questions. For Coursera, MultiFieldQA, and MultiDoc2Dial datasets, we do not further enlarge their sizes to avoid introducing artificial data.

Hereafter, we refer to the expanded dataset as the \textbf{full question set} and the original, pre-expansion dataset as the \textbf{sample question set}.

\subsection{Dataset Statistics}
After expansion, we obtain 19,188 questions, of which 13,651 require context to be answered using the filtering method from \cref{subsec: dataset_filtering}, as listed in Table~\ref{tab:full_datasets}. Notably, questions grounded in factual knowledge, such as those from Coursera, show a high removal rate. Similarly, questions drawn from well-known books or requiring multi-hop reasoning often exhibit a higher likelihood of being directly answered by LLMs without context. Comparing the 12 individual datasets, we observe a similar filtering rate between the sample and the full question sets (see Tables~\ref{tab:benchmark_overview} and~\ref{tab:full_datasets}), indicating that both sets follow a similar distribution.

\begin{table}
    \small
    \resizebox{\columnwidth}{!}{
    \begin{tabular*}{\columnwidth}{l|rrr}
    \toprule
    \textbf{Dataset}
    & \textbf{\# Questions} & \textbf{\# Kept Q} & \textbf{\% Kept Q} \\ \midrule
    Coursera   & 172  & 54   & 32   \\
    NQ         & 1,109 & 373  & 34  \\
    NovelQA    & 2,283 & 869  & 38  \\
    2WikiMHQA  & 2,300 & 1,036 & 45   \\
    HotpotQA   & 2,200 & 1,113 & 51  \\
    MuSiQue   & 2,200 & 1,663 & 78  \\
    MultiFieldQA       & 150  & 121  & 81  \\
    NarrativeQA    & 2,211 & 1,880 & 85 \\
    QASPER     & 2,718 & 2,674 & 98  \\
    QuALTY    & 2,725 & 2,725 & 100 \\
    TOEFL-QA   & 962  & 962  & 100 \\
    MultiDoc2Dial    & 158  & 158  & 100 \\ \midrule
    \textbf{Total}     & \textbf{19,188} & \textbf{13,628} & \textbf{71}\\ \bottomrule
    \end{tabular*}}
    \caption{Statistics of the \textit{full question set}, ordered by increasing percentage of questions kept after filtering out questions needing no context.}
    \label{tab:full_datasets}
\end{table}

\section{Evaluation Methodology} 
\subsection{Evaluation Framework}
Our evaluation of RAG and LC is conducted in the following three phases.

Phase 1: Empirical Study on Retrievers. 
We evaluate five retrievers: BM25, Contriever, OpenAI Embeddings, Llama-Index, and RAPTOR, on the sample question set. The retriever yielding the best performance is then selected for subsequent comparisons with LC on the full question set.

Phase 2: Comparing RAG and LC. 
Using the best retriever, RAG is compared with LC by answering questions on the full question set. Both methods use the same underlying LLM for question answering. For RAG, relevant documents or chunks are fetched from the available context and provided to the LLM as input to generate answers. In contrast, for LC, the entire context available to the question is given to the LLM, with truncation from the back of the context applied if the context exceeds the model's context window. The evaluation metrics are explained in \cref{sub:evaluation}.

Phase 3: In-depth Analysis. 
We focus on 4 specific subsets of questions: 1) those answered correctly only by RAG, 2) those answered correctly only by LC, 3) those RAG gives better answers, and 4) those LC gives better answers. These subsets are analyzed to understand the types of questions each method excels at, providing insights into the strengths and limitations of both approaches in different scenarios.

\subsection{Retriever Selection} \label{sub:retrieval_methods}

Figure~\ref{fig:timelines} shows that existing studies primarily select one or more chunk-based retrieval methods, while index- and summarization-based retrievers are less frequently evaluated. In our study, we evaluate various retrieval methods to ensure that RAG is supported by the most effective retrievers.

For chunk-based retrieval, we use BM25~\cite{robertson2009probabilistic}, Contriever~\cite{izacard2021unsupervised}, and OpenAI's text-embedding-3-Small. BM25 serves as a classic baseline, while Contriever and text-embedding-3-Small represent embeddings from well-performing closed-source and open-source models, respectively. 

For index-based retrieval, we employ Llama-index and leverage two indexing methods that suit long documents. Specifically, \textbf{tree-index} organizes documents into a hierarchical tree structure, enabling efficient retrieval of context. The root node contains a high-level summary, while subsequent child nodes store progressively finer-grained representations. When queried, the retrieval process navigates through this hierarchy, starting from the top-level summary and moving down to more specific nodes as needed. \textbf{Sentence Window Retriever} focuses on local, sentence-level context rather than entire documents or large text chunks. It creates smaller “windows” of a few sentences each. When a query arrives, the retriever searches these windows to identify segments most semantically similar to the query. By working at a finer granularity, the sentence window retriever provides more targeted and contextually accurate snippets of text, improving the model’s ability to answer specific questions.

For \textbf{summarization-based} retrieval, we use RAPTOR~\cite{sarthi2024raptor}. It constructs a hierarchical tree by recursively clustering text chunks based on semantic similarity, summarizing each cluster into a parent node, and continuing this process until no further clustering is possible. After constructing the tree, we apply the collapsed tree traversal approach, as previous work has demonstrated its superior performance. This approach flattens the hierarchical structure into a single layer and compares the query against all nodes across every level simultaneously. The top-k most relevant nodes are then selected based on a predefined token limit, ensuring that the retrieved information maintains the appropriate level of granularity.

Although RAPTOR’s implementation appears similar to the Llama Tree Index, they differ in both construction and navigation. First, Llama Tree Index groups consecutive nodes, while RAPTOR freely clusters nodes from far positions, and even allows a single node to appear in multiple clusters. Second, Llama Tree Index navigates down the hierarchy to retrieve only leaf nodes, while RAPTOR evaluates all nodes from all layers simultaneously. Hence, RAPTOR can retrieve not only original texts but also generated summaries.

\subsection{Evaluation Metric} \label{sub:evaluation}

\begin{figure}
\centering
\includegraphics[width=0.8\linewidth]{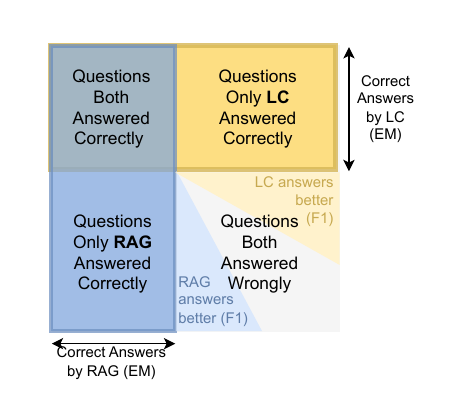}
\caption{Evaluation Matrix for In-depth Analysis.}\label{fig:eval}
\end{figure}

\begin{table*}
    \small
    \resizebox{\textwidth}{!}{
    \begin{tabular*}{\textwidth}{l|r|rr|rr|rr}
    \toprule
    \textbf{Dataset} & \textbf{\# Questions} & \textbf{LC Correct} & \textbf{RAG Correct} & \textbf{LC Only} & \textbf{RAG Only} & \textbf{LC Better} & \textbf{RAG Better} \\
    \midrule
    Coursera & 54 & 26 & 20 & 10 & 4 & 10 & 4 \\
    2WikiMHQA & 1,036 & 594 & 431 & 242 & 79 & 265 & 107 \\
    HotpotQA & 1,113 & 876 & 723 & 212 & 59 & 231 & 67 \\
    MultiFieldQA & 121 & 63 & 60 & 14 & 11 & 44 & 21 \\
    NQ & 373 & 189 & 138 & 75 & 24 & 104 & 35 \\
    NarrativeQA & 1,880 & 558 & 405 & 276 & 123 & 685 & 281 \\
    QASPER & 2,674 & 884 & 863 & 517 & 496 & 1,011 & 762 \\
    QuALITY & 2,725 & 2,290 & 2,050 & 402 & 162 & 402 & 162 \\
    TOEFL-QA & 962 & 895 & 884 & 26 & 15 & 26 & 15 \\
    MuiQue & 1,663 & 821 & 663 & 344 & 186 & 426 & 225 \\
    MultiDoc2Dial & 158 & 14 & 38 & 5 & 29 & 65 & 58 \\
    NovelQA & 869 & 466 & 408 & 164 & 106 & 164 & 106 \\
    \midrule
    \textbf{Overall} & \textbf{13,628} & \textbf{7676} & \textbf{6,683} & \textbf{2,287} & \textbf{1,294} & \textbf{3,433} & \textbf{1,843} \\
    \bottomrule
    \end{tabular*}}
    \caption{Performance of LC and RAG across different datasets. We report the number of questions answered correctly by each method, as well as the breakdown of questions where: only LC answers correctly (LC Only), only RAG answers correctly (RAG Only), LC outperforms RAG (LC Better), and RAG outperforms LC (RAG Better).}
    \label{tab:long_rag_results}
\end{table*}

We use a win-lose rate system to compare LC and RAG, as illustrated in Figure~\ref{fig:eval}. The horizontal yellow block represents the questions that the LLM answers correctly using LC, while the vertical blue block represents the questions that the LLM answers correctly using RAG. Their overlap in the top-left corner represents the questions that both methods answer correctly. We apply an \textit{Exact Match} (EM) score strictly to all questions to determine the correctness of the answers. Excluding the overlap, the top right block indicates the questions that \textbf{only LC} answers correctly, and similarly, the bottom left block indicates the questions that \textbf{only RAG} answers correctly.

The remaining gray block represents the questions that both RAG and LC answer incorrectly, as judged by Exact Match. Since many questions involve long open-ended responses, we calculate the $F1$ scores of the answers provided by both methods against the ground truth. If RAG achieves a higher $F1$ score than LC, we consider RAG to have answered the question better, and vice versa for LC. A detailed explanation of $F1$ score calculation is provided in \cref{app:f1}

The loose evaluation setting considers all cases in which one method outperforms the other, including 1) when one method obtains the correct answer and the other is wrong under EM, and 2) when one method achieves a higher $F1$ score. We adopt this loose evaluation because references for some datasets are long, open-ended answers, making it very unlikely to match them exactly under EM. In addition, some short answers (about 5–6 words) may differ slightly from the reference while still conveying the correct idea. Although these answers would be marked incorrect by EM, they might attain a high $F1$ score. Hence, comparing $F1$ scores helps compensate for the strictness of EM.
\section{Experiments}
To obtain answers, we use the same prompt \textit{``From the context: [context], answer the questions briefly with no explanation.''} for both retrieval and long context settings. For MCQ questions, we add one sentence \textit{``Answer the question with the letters of the correct options (e.g. A, BC, C, ACD, etc.) without including text''}. These prompts ensure LLMs to directly answer the questions, which makes evaluation more convenient.

\subsection{Phase 1: Retrievers}
Evaluated on the sample question set, Table~\ref{tab:retrieval_methods} reports the results of chunk-, index-, and summarization-based retrievers. Among them, RAPTOR performs the best with a correct answer rate of 38.5\%, while Index-based retrievers outperform chunk-based retrievers. Within index-based retrievers, the “RAG Only” score for Tree Index is much lower than that for Window Parsing (82 vs. 91), and their “RAG Better” scores are nearly identical (234 vs. 237). This discrepancy suggests that Tree Index may be undervalued in the “RAG Only” metric but still contributes in open question scenarios that require long answers.

We further observe the questions and contexts that each retriever exclusively answers correctly. RAPTOR shows stronger ability than other retrievers, especially in scenarios that require an entire understanding of the document, like research papers. Chunk-based methods struggle when required information is spread across multiple chunks. Index-based retrievers are not as strong in overall understanding as RAPTOR, but they show good ability in interpreting dialogues. Therefore, we select RAPTOR as the primary retriever for evaluation on the full question set.

\begin{table}
    \centering
    \resizebox{\columnwidth}{!}{
    \begin{tabular}{l|lccc}
    \toprule
    \textbf{Type}   & \textbf{Retriever}    & \textbf{Correct (\%)} & \textbf{RAG Only} & \textbf{RAG Better} \\ \midrule
    \multirow{3}{*}{Chunk} 
                                & BM25    & 319 (20.4)  & 50   & 141  \\ 
                                & Contriever & 315 (20.1)  & 43    & 143 \\ 
                                & Text-emb-3-small  & 338 (21.6) & 47   & 151 \\ \midrule
    \multirow{2}{*}{Index} 
                                & Tree Index  & 470 (30.1) & 82 & 234  \\ 
                                & Window Parsing  & 555 (35.5) & 91 & 237 \\ \midrule
    \multirow{1}{*}{Summarization} 
                                & RAPTOR  & \textbf{602 (38.5)} & 97 & 258 \\ \bottomrule 
             \end{tabular}}
        \caption{Comparison of different retrieval methods}
    \label{tab:retrieval_methods}
\end{table}

\subsection{Phase 2: Comparing LC and RAG}

We compare LC and RAG on the filtered, full question set. The results across 12 datasets are summarized in Table~\ref{tab:long_rag_results}. Overall, LC correctly answers 56.3\% of the questions, while RAG provides correct answers to 49.0\%. LC correctly answers more than 2,000 questions that RAG misses, while RAG exclusively answers almost 1,300 questions. When looking at the loose evaluation setting, LC answers 3,433 questions better than RAG, and RAG answers 1,843 questions better than LC. The gap further widens compared to strict setting, indicating long-context LLM's ability to answer questions with open long answers is also strong. 

Looking at individual datasets, in MultiDoc2Dial, RAG exhibits better performance than LC in strict evaluation (5 vs 29), but is surpassed by LC in loose evaluation (65 vs 58). In contrast, on datasets like NarrativeQA and QuaLITY, LC shows a strong lead not just in overall correctness but also in the number of questions that are answered better. 
Collectively, the results show that both methods have unique strengths and limitations.

Although LC shows better overall results than RAG, out of the 13,628 questions, almost 10\% can be only answered correctly by RAG, which is not a small ratio. This shows that retrievers cannot be simply replaced by long-context LLM in searching. This also motivates us to further examine what kind of questions (and context) can be only answered correctly by RAG (or LC).

\subsection{Phase 3: In-Depth Analysis} \label{subsec: in-detail}
\begin{figure}[t]
\includegraphics[width=\linewidth]{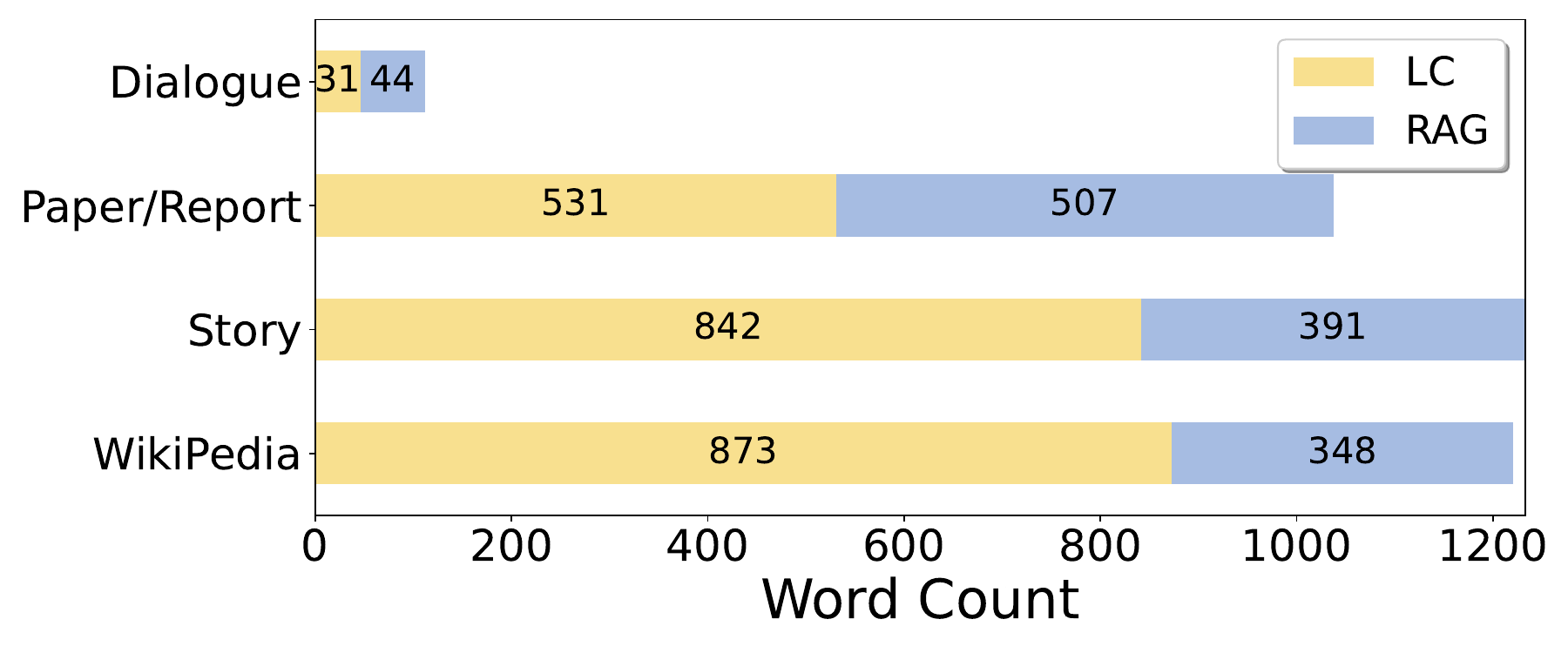}
\caption{Performance breakdown by knowledge source for LC Only and RAG Only.}\label{fig:knowledge_source_only}
\end{figure}

\begin{figure}[t]
\includegraphics[width=\linewidth]{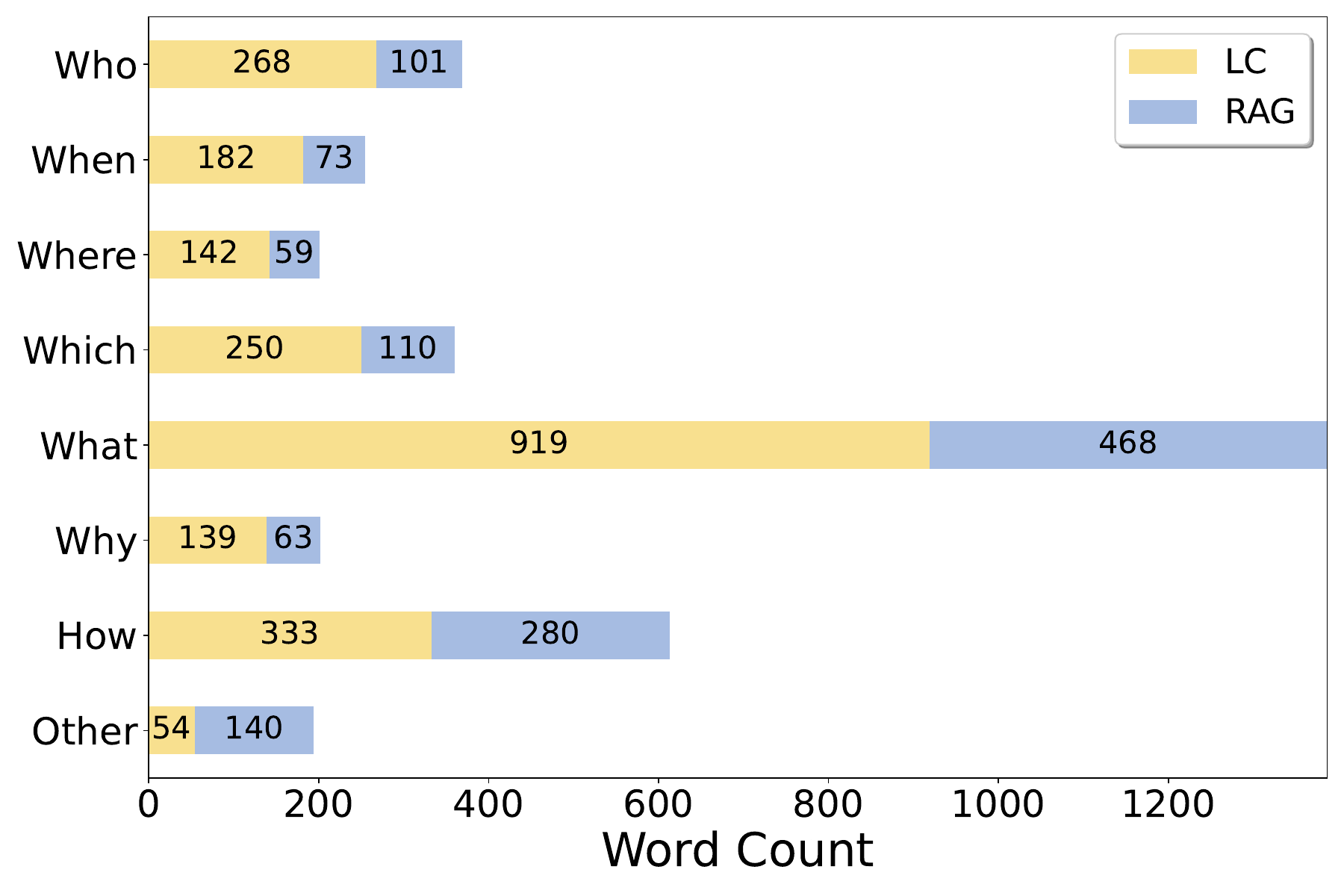}
\caption{Performance breakdown by question type for LC Only and RAG Only.}\label{fig:question_type_only}
\end{figure}

The overall results are influenced by the combined effects of different scenarios, so we need to separately analyze each scenario to see if more detailed results can be obtained. We analyze the performance of LC and RAG across different knowledge sources (Figure~\ref{fig:knowledge_source_only}) and question types (Figures~\ref{fig:question_type_only}). Here, we use EM Scores only, for a strict evaluation standard. We also report the results for loose evaluation standard (i.e., EM Scores and $F1$ Scores) in \cref{sec:appendix}, which shows similar trends.

From Figure \ref{fig:knowledge_source_only}, it is evident that LC excels with knowledge sources such as Wikipedia and stories. However, the Wikipedia context is collected by adding extensive noise to create long context, which generally makes the context less relevant to the question, with only a small portion being useful. This synthetic context formation partially simulates the RAG process and may introduce an unfair bias against the RAG pipeline. In addition, summarization-based retrieval methods may split Wikipedia articles unnaturally, generating less meaningful summaries. LC’s strong performance demonstrates that long-context LLMs are robust to noise in such forms of context.

In contrast, RAG performs better with dialogue-related sources and achieves comparable performance with papers or reports. The information in these sources is naturally segmented, conversations have turns, and papers and reports have clearly defined sections or subsections, making the retrieval of key segments easier.

Figure \ref{fig:question_type_only} shows that LC performs better for fact-based questions such as ``Who'', ``Where'', and ``Which''. These questions often benefit from having all the relevant context available in a dense region close to the answer. RAG, however, is largely comparable to LC for more open-ended questions such as ``How'', which often require synthesizing information from multiple sources and therefore benefit from retrieval-based approaches.

Furthermore, RAG outperforms LC in the ``Other'' questions, which consist mainly of general questions that can be answered with ``Yes'' or ``No''. We hypothesize that the reason could be due to the training data. Long-context LLMs are more familiar with phrasing of common type questions than general questions. Words like ``Who'' or ``Where'' act as keywords for long-context LLMs to search, while retrievers use these keywords not so well. 


\subsection{Word Frequency Visualization}

\begin{figure}[t]
\includegraphics[width=\linewidth]{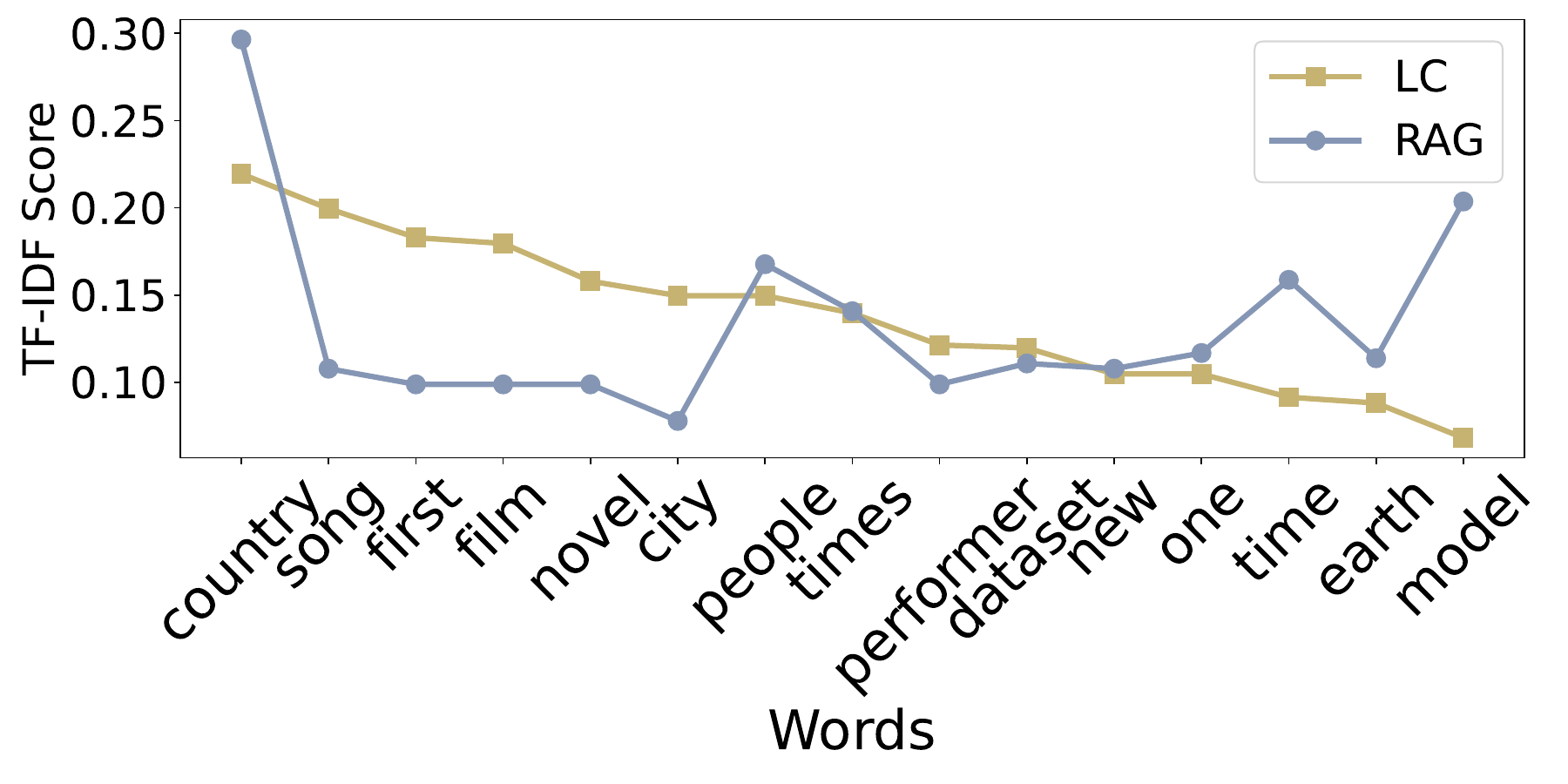}
\caption{Top 15 Words based on TF-IDF Score for LC Only vs. RAG Only.}
\label{fig:word_frequency_line_chart}
\end{figure}
To better understand the scenarios that LC and RAG each excels at, we visualize the word frequencies by their TF-IDF scores, plotted in Figure \ref{fig:word_frequency_line_chart}. The TF-IDF scores were calculated from questions in the datasets where either LC or RAG produced correct answers exclusively. Specifically, all questions from each dataset are concatenated and treated as a single document for this analysis, meaning that the TF-IDF scores primarily reflect the term frequency within each dataset. Stopwords are removed and not shown in the plot. 

Figure \ref{fig:word_frequency_line_chart} presents the top 15 words  that appear most frequently combined in both LC only and RAG only questions. Words such as `song', `film', and `novel' have higher TF-IDF scores for LC, suggesting that LC performs better with narrative topics. Conversely, words like `country', `dataset', and `model' have higher scores for RAG, indicating its strength in retrieving information on technical or data-oriented topics. This analysis underscores the complementary strengths and limitations of LC and RAG in handling different types of questions.

\subsection{Impact of Generation Model in RAG} We now evaluate the impact of different generation models on RAG's performance. Table \ref{tab:generation_models} shows the results of using GPT-4o and GPT-4-Turbo as the generator with three retrievers (BM25, Tree Index, RAPTOR), each of which represents one retriever type. The results indicate that the performance of different generation models remains largely consistent regardless of the retriever used. RAPTOR performs the best across both generation models, though there is a slight decrease in performance when using GPT-4-Turbo compared to GPT-4o.

While GPT-4o slightly outperforms GPT-4-Turbo across all retrievers, the differences are marginal. This implies that both generation models are capable of generating high-quality responses, and the choice between them may depend more on other factors such as efficiency or resource availability. The consistency across retrievers also demonstrates that the retrieval method plays a larger role in determining overall performance than the specific generation model used. We will report the results from other models and the experiment is in progress.

\begin{table}
    \centering
    \resizebox{\columnwidth}{!}{
    \begin{tabular}{l|lccc}
    \toprule
    \textbf{Retriever}   & \textbf{Model}    & \textbf{Correct (\%)} & \textbf{RAG Only} & \textbf{RAG Better} \\ \midrule
    \multirow{2}{*}{BM25} 
                                & GPT-4o    & 319 (20.4)  & 50  & 141  \\ 
                                & GPT-4-Turbo  & 310 (19.8) &  51  & 152 \\ \midrule
    \multirow{2}{*}{Tree-Index} 
                                & GPT-4o  & 470 (30.1) & 82 & 234  \\ 
                                & GPT-4-Turbo  & 458 (29.3) & 81 & 229 \\ \midrule
    \multirow{2}{*}{RAPTOR} 
                                & GPT-4o  & \textbf{602 (38.5)} & 97 & 258 \\
                                & GPT-4-Turbo  & \textbf{589 (37.7)} & 99 & 295 \\ \bottomrule 
             \end{tabular}}
        \caption{Results of using different generation models}
    \label{tab:generation_models}
\end{table}

\subsection{Case Study}\label{sec:case_study}
For a deeper understanding of the difference between LC and RAG, we conduct a case study to analyze the frequent errors from each method, and present them in Tables \ref{tab:case_study1} and  \ref{tab:case_study2}. We manually examine the questions that  only RAG made mistakes, and those only LC made mistakes. 

\begin{table}
\small
    \centering
    \begin{tabular}{>{\raggedright\arraybackslash\tt}p{0.46\textwidth}<{}}
    \hline
    \textbf{Question}: What is the debt-to-GDP ratio of the country where Anthony Upko was formerly involved in the government?\\
    \textbf{Wrong Answer}: The context does not provide the debt-to-GDP ratio for Nigeria.\\
    \textbf{Gold}: 11 percent \\
    \textbf{Relevant Sents}: 1. Nigeria is the world's 20th largest economy ...  the debt-to-GDP ratio is only 11 percent. 2. Anthony Ukpo was Minister of Information and Culture, and then Governor of Rivers State, Nigeria.\\\hline
    \textbf{Question}: When is the performer of song Swing Down Sweet Chariot 's birthday?\\
    \textbf{Wrong Answer}: May 8, 1940	\\
    \textbf{Gold}: January 8, 1935\\
    \textbf{Relevant Sents}: 1. Swing Down Sweet Chariot is a traditional song ... recorded by Elvis Presley. \\2.  Elvis Aaron Presley (January 8, 1935 – August 16, 1977), also known as ... \\\hline
  \end{tabular}
  \caption{Examples cases where RAG made mistakes}
  \label{tab:case_study1}
\end{table}

The most frequent mistake made by RAG is its failure to retrieve the relevant context, leading to its refusal to answer the question. As shown in Table \ref{tab:case_study1}, the model correctly identifies that Anthony Upko was formerly involved in the government of Nigeria but fails to retrieve the debt-to-GDP ratio as part of the context. This retrieval failure can arise due to two possible reasons: the retriever might fail to locate the relevant sentences from documents, or the sentences may be split across two chunks, with the debt-to-GDP ratio lacking a clear subject. Interestingly, when provided with the same prompt, LC rarely reports a lack of context, suggesting its robustness in handling such cases.

Another error made by RAG is misinterpreting partial context. In the second example, where RAG incorrectly answered the birthday, the model retrieved May 8, 1940, instead of the correct date, January 8, 1935. This occurred because the sentence `Swing Down Sweet Chariot is a traditional song ... recorded by Elvis Presley' spans too long, creating ambiguity in linking the birthday to the correct person. This type of retrieval failure highlights a core limitation: RAG relies heavily on retrieving continuous text spans, and any fragmentation or overly long context can lead to an incomplete understanding. In contrast, LC tends to provide more holistic answers when processing longer contexts directly, as it bypasses the dependency on a retrieval module.

\begin{table}
\small
    \centering
    \begin{tabular}{>{\raggedright\arraybackslash\tt}p{0.46\textwidth}<{}}
    \hline
    \textbf{Question}: Do the tweets come from a specific region?\\
    \textbf{Wrong Answer}: Yes, the tweets come from 16 different countries.\\
    \textbf{Gold}: No\\
    \textbf{Relevant Sents}: This helped us narrow down our query space to 16 countries.\\\hline
    \textbf{Question}: Where did Valancourt lose his wealth?\\
    \textbf{Wrong Answer}: In Gambling.\\
    \textbf{Gold}: Paris\\
    \textbf{Relevant Sents}: Returning to her aunt's estate, Emily learns that Valancourt has gone to Paris and lost his wealth.\\\hline
  \end{tabular}
  \caption{Examples representing common cases where only RAG answers correctly}
  \label{tab:case_study2}
\end{table}

Wrong answers by LC are often caused by question misinterpretation. For instance, as shown in Table~\ref{tab:case_study2}, when asked whether the tweets come from a specific region, LC answers `yes', referencing that the tweets originate from 16 countries. It fails to interpret the relationship between `a specific region' and `16 different countries'. In another example, when asked `where' Valancourt lost his wealth, the model identifies the correct sentence but answers `how' instead of `where'. These examples highlight that LC sometimes struggles to align its semantic understanding with the required level of specificity or perspective, resulting in answers that are related but not addressing the question's intent. In both cases, the LLMs are able to locate the related texts from the documents, but the reasoning ability might be affected by the noise.
\section{Discussion}
\label{sec:discussion}

\subsection{What is Long Context?}
Although we have reviewed 9 studies that either directly or implicitly compare or integrate RAG and Long Context, very few studies clearly define what Long Context is. To this end, we separately interpret the two words `long' and `context'.

\paragraph{Long.}
Out of the 9 studies reviewed earlier, only 2 studies, ChatQA2 and LongBench v2 explicitly define Long Context as greater than 32k and greater than 8k tokens respectively. For other studies, we can only infer their definitions of ``long'' based on the models and datasets they use. It seems that three studies consider 8k as a minimum requirement for long context, and another three studies set this requirement at 16k. Lastly, OP-RAG regards 128k as long context.

In short, each work defines `Long Context' based on its own criteria due to the lack of a clear standard. Moreover, as the context windows of language models continue to expand, the terms `long' and `short' are relative. For example, 4k tokens are not considered `long context' in any of the reviewed studies but are extremely long for BERT-base models, which support only 512 tokens. As a result, the definition of `long' remains ambiguous, leading to inconsistent use of this concept among researchers. In practice, the definition of `long' is complicated, depending on the context length of latest LLMs, and the length of the documents in targeted domain. 

\paragraph{Context}
In the English dictionary, `context' is defined as ``the situation within which something happens, and that can help explain it''. By this definition, the context of a question is expected to ``help explain it'', implying that the context should have strong relevance to the question. However, long-context datasets are not always constructed with this principle in mind. The construction of long-context datasets can generally be categorized into two types:

\textit{Realistic Long Texts}: These datasets originate from sources such as novels, research papers, or other lengthy narratives, exemplified by datasets like NovelQA. Such datasets typically pose challenges that involve reading comprehension and require models to process and synthesize dense information spread across a cohesive, extended text.
 
\textit{Synthetic Long Texts}: These datasets are often created by concatenating smaller, query-relevant segments of text, such as Wikipedia-sourced datasets in LongBench. This construction process may involve stitching together Wikipedia excerpts, injecting noise, or combining unrelated passages to simulate a long document. 

A critical observation is that realistic long contexts align more closely with reading comprehension tasks, where models primarily absorb and reason over information. Such datasets have high contextual relevance, since the questions are normally based on the documents that users provided. In contrast, synthetic long contexts often resemble factual reasoning tasks, where models retrieve and verify knowledge.
Such datasets inherently incorporate a pre-processing step like a RAG pipeline. 
They can assess the impact of information placement on model performance, such as the lost-in-the-middle phenomenon.

On the other hand, realistic and synthetic long texts can only serve as proxies to reflect context relevance to some extent. The scope of the context is question-dependent and difficult to define clearly.

\subsection{How to Compare or Combine LC \& RAG?} \label{subsec: discussion2}

The lack of a clear definition for long context also indicates the absence of a coherent framework for comparing or combining LC and RAG. We propose such a framework by examining three key perspectives: context length, context relevance, and experiment design.

\paragraph{Context Length.}
From the model’s perspective, context length refers to the maximum number of tokens a model can process. From the dataset’s perspective, it denotes the amount of text provided with a question. In synthetic datasets, context length is flexible, but this introduces a trade-off between length and relevance. Adding irrelevant information as context may help to test a model's robustness to noise, but such testing may not represent real-world use cases. Therefore, any framework for comparing LC and RAG should clearly define what is considered `long', while indicating whether this length criterion originates from the model's capabilities, the dataset's design, or both.

\paragraph{Context Relevance.}
An evaluation framework must also address the relevance of the text provided as input to the model. It is crucial to distinguish between realistic long contexts and synthetic long contexts. When benchmarks include both types, separate evaluations are necessary, as synthetic contexts often have low relevance and may not accurately reflect real-world scenarios.

Interestingly, the construction of synthetic long contexts often mirrors RAG pipelines. Providing an entire curated text to an LLM as context essentially represents a `long context RAG' approach, given that such text is assembled during dataset creation. Further chunking can introduce biases against RAG by disrupting the continuity of information within each piece.

Additionally, many benchmarks categorize tasks as `single-doc' or `multi-doc' based on whether the text originates from a single source or multiple documents. While convenient, this categorization does not perfectly align with `realistic' or `synthetic' contexts. A single document may sometimes be artificially composed of smaller fragments, while a multi-sourced document might involve highly relevant sources, such as a group of research papers discussing the same problem.

The key issue remains determining to what extent the context provided as input to LLMs contains sufficient and relevant content to answer the question, without introducing unnecessary or unrelated information.

\paragraph{Experiment Settings.}
When investigating LC and RAG, the experimental objectives can be broadly grouped into two categories: comparison and combination.

\textit{Short RAG v.s. Long Single Input}: one might compare a short-context RAG pipeline against a long-context single-input setup, analyzing both performance and computational cost. This provides insights into the trade-off between running an extra retrieval pipeline for shorter contexts versus allowing the model to process a larger uninterrupted text.

\textit{Long RAG v.s. Long Single Input}: One may also compare a long-context RAG pipeline with a long-context single-input approach. Here, the goal is to see whether chunking or filtering more relevant content through retrieval can outperform or complement a fully integrated long-context approach by truncating exceptionally long documents.

In the first setting, the retrieval pipeline naturally reduces the number of tokens. In the second setting, the context length remains the same for both methods, with the only difference being how the text is processed.

\textit{RAG over Increasing Context}: Another possible goal is understanding how RAG performance changes with increasing context lengths. In this scenario, the ``LC'' refers specifically to how many tokens a model can handle. This line of work can reveal how well RAG pipelines scale when models absorb increasingly larger inputs.

On the other hand, findings from evaluations often serve as guidelines for settings that address real-world problems. In this sense, RAG and LC may complement each other in real-world settings, depending on the characteristics of the data source and the types of questions to be answered.

\subsection{Revisiting All Studies}
Based on the earlier discussion, the exploration of LC and RAG methods in LLMs highlights some critical challenges that researchers often overlook.

\paragraph{Trade-off between Context Length and Relevance.}
Many studies hesitate between using flexible synthetic context with noisy concatenated contexts, or realistic context with dense information but less availability. Among the 9 studies, 6 select synthetic context as part of the datasets. Our own evaluation has also selected synthetic context datasets, but we consider the influence of synthetic long context and separately evaluate their results by context source; e.g. a Wikipedia source with manually added noises represents low context relevance.

Several studies have attempted to address this challenge. LongBench recently updated v2 which collects only realistic data. Despite a smaller scale, LongBench v2 shows substantial improvement in context relevance compared to its first version.
LongRAG retrieves from a massive corpus for all questions, instead of assigning one context to each question. This method avoids retrieving from a synthetic long context and is hence recommendable. 

\paragraph{Diversity in Retrieval Mechanisms.}
In the comparison of RAG and LC, RAG is often under-represented due to an over-reliance on traditional retrieval strategies. Among the 9 studies, 5 experiment with different retrievers, only 2 try different chunking sizes, and none consider any retrieval method beyond chunk-based retrievers. Although we experiment with index-based and summarization-based retrievers, we cannot promise that our selected method outperforms all retrieval strategies.

For investigating RAG performance over increasing context, some studies propose their own strategies for chunking and placing RAG. OP-RAG proposes preserving the original order of chunks from the context, while LC LLM-RAG proposes placing higher-scored chunks at the front and back. In addition to more advanced retrievers, certain information retrieval (IR) ~\cite{ir} techniques like relevance feedback~\cite{relevance_feedback} or query expansion~\cite{query_expansion} might further enhance RAG performance, yet these have been overlooked in existing frameworks.

\paragraph{Computational Cost.}
Most existing studies test on 6 to 8 datasets, and it becomes increasingly expensive to conduct experiments on too many models. This is especially the case when new long-context LLMs are being released at a very fast pace. Hence, any work might be questioned because the experiment results are only applicable to one or a few models. Among all works, LC RAG Performance includes the largest number of models (20). While their efforts are remarkable, they only experiment on 3 datasets. FinanceBench~\cite{financebench} looks at finance domain, Databricks DocsQA is based on Databricks platform, and NQ as shown table \ref{tab:benchmark_overview} as a very low rate of requiring external knowledge. This is not meant as criticism but rather to show the trade-off between testing many models and having a comprehensive benchmark.
\section{Conclusion}
In this paper, we survey existing studies comparing or combining LC and RAG, analyzing why different implementations may result in some conflicts among their insights.
Therefore, we present a thorough comparison of LC and RAG approaches by leveraging a diverse set of long context QA datasets. We filtered out questions that could be answered from parametric knowledge, ensuring a fair comparison by focusing on questions that required external context. Along these lines, we have developed a systematic filtering and evaluation process, identified the best retrieval method, and expanded the dataset to provide a statistically significant basis for analysis. The results indicate that LC generally outperforms RAG for tasks involving well-structured, dense contexts—such as Wikipedia articles and books—and is better at answering questions requiring specific information.
By contrast, RAG demonstrates advantages in handling fragmented information, particularly in dialogue-based scenarios and for more general questions.

Beyond merely presenting the experimental results and findings, we delve deeper into the concept of long context and examine how LC and RAG should be compared. Our discussion aims to ensure that the insights gained are more impactful and applicable to real-world scenarios.

\section*{Limitations}
While our study provides valuable insights into the comparative strengths and weaknesses of Long Context (LC) and Retrieval-Augmented Generation (RAG) approaches, it is important to acknowledge three limitations that may impact the generalizability and comprehensiveness of the findings:

Our analysis is limited to text-based long contexts, and neglecting other modalities such as audio, video, or multi-modal contexts. The applicability of these insights to non-textual long-context scenarios remains unexplored, which may limit the broader applicability of the findings to multi-modal applications.

Our work focuses on existing papers that compare and combine RAG with long-context LLMs. Therefore, we mainly survey the retrievers and LLMs used in those papers, rather than all available retrievers and long-context LLMs.

Our experiments rely on existing LC and RAG implementations, including specific retrieval methods and strong long-context models. As the field continues to evolve, newer models or retrieval strategies may alter the comparative outcomes. However, our evaluation framework is still applicable to future evaluation.

\section*{Ethical Considerations}
Advanced Long Context LLMs equipped with strong RAG capabilities could be misused to generate misleading or harmful content, such as fake news or propaganda. Their long-context capability could amplify the scale and believability of such content. Researchers should prioritize safety and transparency in model usage to mitigate the risk.

\bibliography{anthology,main}
\appendix
\newpage
\section{F1 Score Computation} \label{app:f1}
To calculate the $F1$ score, we first convert both the prediction and the reference text into sets of unique tokens. Tokens appearing in both sets count as true positives (TP), tokens present only in the prediction are 
false positives (FP), and tokens missing from the prediction but in the 
reference are false negatives (FN). Precision is defined as 
\(\tfrac{\text{TP}}{\text{TP} + \text{FP}}\), recall as 
\(\tfrac{\text{TP}}{\text{TP} + \text{FN}}\), and the F1 score is their 
harmonic mean:

\[
F1 = 2 \times 
\frac{\text{precision} \times \text{recall}}{\text{precision} + \text{recall}}.
\]

\noindent \textbf{Example:}

\begin{quote}
"cat leaps table quickly"(prediction)  \\
"the cat leaps over the table" (reference)
\end{quote}

\noindent The corresponding sets are:
\[
\text{prediction\_set} = 
\{cat, leaps, table, quickly\}
\]
\[
\text{gold\_set} = 
\{the, cat, leaps, over, table\}.
\]

\noindent Here, 
\(\{\texttt{cat}, \texttt{leaps}, \texttt{table}\}\) are TP \(= 3\), 
\(\{\texttt{quickly}\}\) is FP \(= 1\), and 
\(\{\texttt{the}, \texttt{over}\}\) are FN \(= 2\). Hence:
\[
\text{precision} = \frac{3}{3 + 1} = 0.75, \quad
\text{recall} = \frac{3}{3 + 2} = 0.60,
\]
\[
F1 = 2 \times \frac{0.75 \times 0.60}{0.75 + 0.60} 
= 0.67.
\]

\section{In-detail Analysis on Loose Evaluation Settings }
\label{sec:appendix}

As a complement to \cref{subsec: in-detail}, we provide a detailed comparison of the performance of LC and RAG under the loose evaluation settings based on Exact Match (EM) and F1 scores.
\begin{figure}[t]
\includegraphics[width=\linewidth]{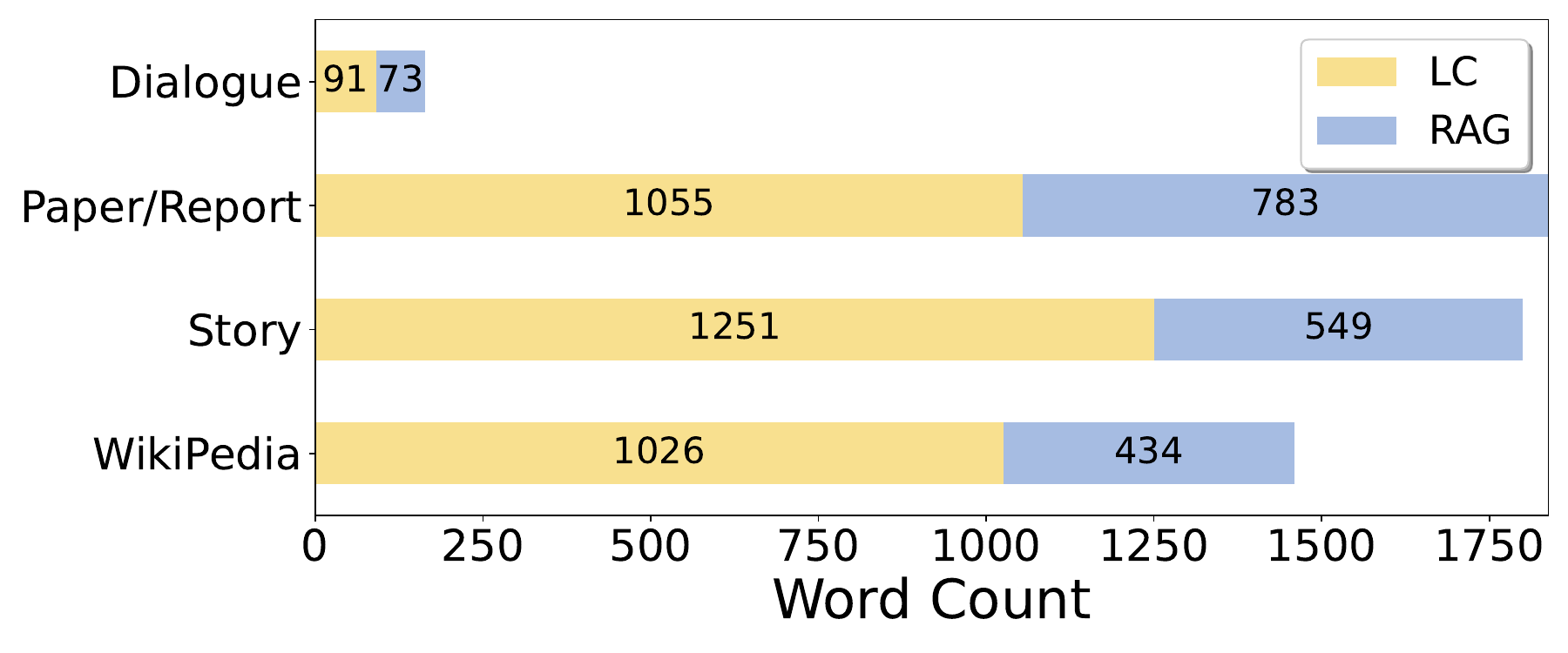}
\caption{Performance breakdown by knowledge source for  LC Better and RAG Better.}
\label{fig:knowledge_source_better}
\end{figure}

\begin{figure}[ht]
\includegraphics[width=\linewidth]{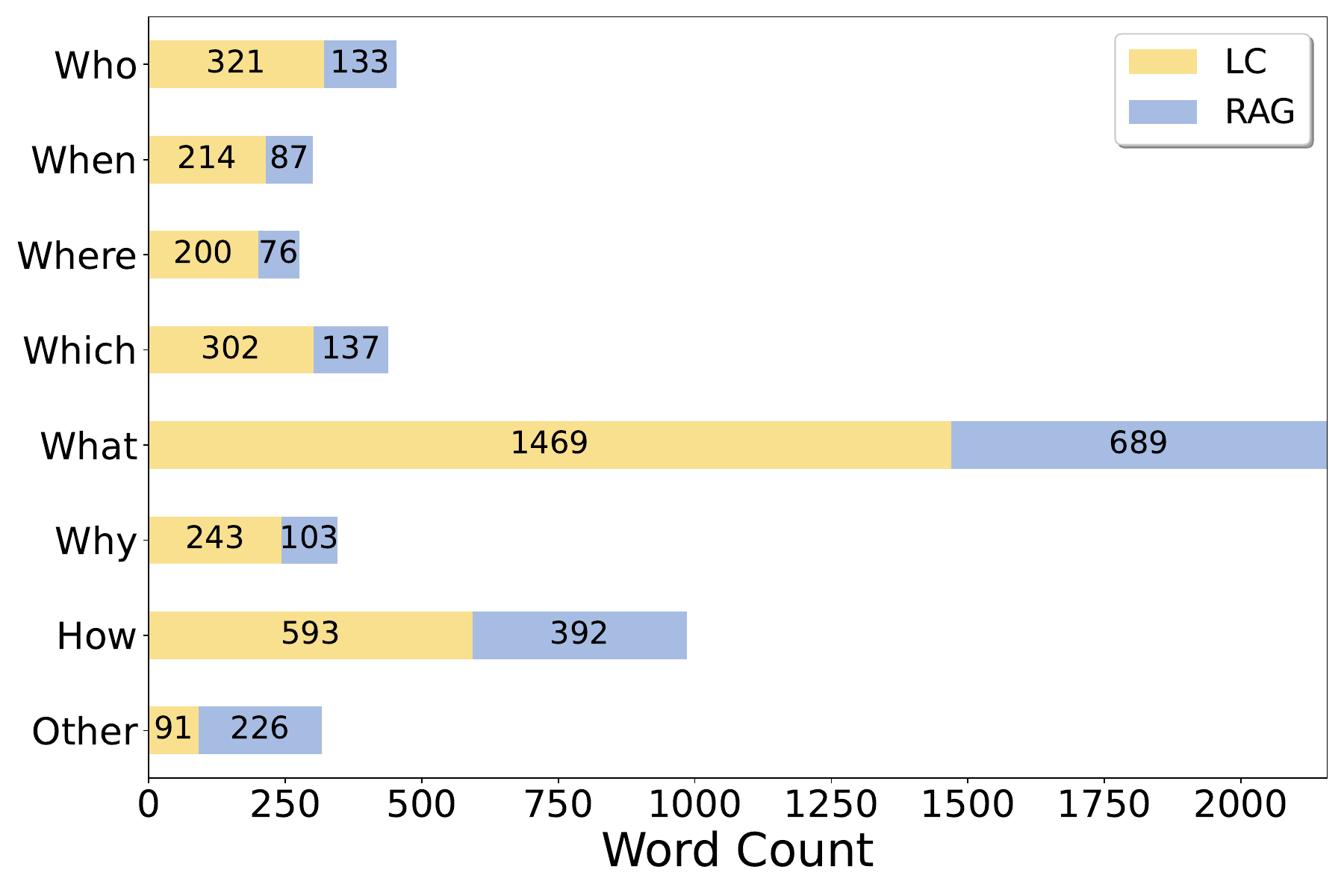}
\caption{Performance breakdown by question type for  LC Better and RAG Better.}\label{fig:question_type_better}
\end{figure}

As shown in Figure \ref{fig:knowledge_source_better}, loose evaluation setting reveals similar trends to the strict setting in the performance of LC and RAG on different knowledge sources. LC outperforms RAG for structured sources like Wikipedia, course websites, and papers/reports, where having complete context is advantageous. This trend is consistent in both evaluation settings. However, RAG performs better with dialogue-based and story-based knowledge sources, where the information is fragmented. The loose evaluation, with the inclusion of F1 scores, shows a slight improvement for RAG in these cases, as partial answers are rewarded more, but the overall trend remains the same.

Figure \ref{fig:question_type_better} highlights the performance of LC and RAG across different question types. For fact-based questions (e.g., ``Who'', ``Where'', ``Which''), LC continues to outperform RAG in both evaluation settings, as these questions benefit from having complete, uninterrupted context. For open-ended questions (e.g., ``How'', ``Why''), RAG shows comparable performance to LC in both settings. The loose evaluation, however, slightly favors RAG due to its ability to synthesize information from multiple sources, as F1 scoring acknowledges partial correctness. In the case of "Other" questions (simple "Yes" or "No" questions), RAG significantly outperforms LC in both evaluation settings, but the advantage is more pronounced in the loose evaluation. The inclusion of F1 scores helps RAG capture partial successes that would be penalized under strict EM-only scoring.

Overall, the figures illustrate that the performance patterns of LC and RAG remain largely consistent across both strict and loose evaluation settings. The key difference is that RAG gains a slight performance boost in the loose evaluation.

\end{document}